\documentclass[11pt,a4paper]{article}
\usepackage[hyperref]{naaclhlt2019}
\usepackage{times}
\usepackage{latexsym}
\usepackage{amsmath}
\usepackage{amsfonts}
\usepackage{graphicx}
\usepackage{booktabs}
\usepackage{amssymb}
\usepackage{url}

\aclfinalcopy 

\newcommand\blfootnote[1]{%
  \begingroup
  \renewcommand\thefootnote{}\footnote{#1}%
  \addtocounter{footnote}{-1}%
  \endgroup
}

\newcounter{notecounter}

\newcommand{\enoteson}{\long\gdef\enote##1##2{{
\stepcounter{notecounter}
\large\bf
\hspace{1cm}\arabic{notecounter} $<<<$ ##1: ##2
$>>>$\hspace{1cm}}}}
\enoteson

\title{Neural Semi-Markov Conditional Random Fields for \\Robust Character-Based Part-of-Speech Tagging}

\author{Apostolos Kemos$^{2*}$ \\\\\\\\
   \\\And
  Heike Adel$^{1,3*}$\\
  $^1$ Center for Information and Language Processing (CIS), LMU Munich, Germany\\
   $^2$ Department of Computer Engineering and Informatics, University of Patras, Greece\\
  $^3$ Bosch Center for Artificial Intelligence (BCAI), Renningen, Germany\\
  {\tt kemos@ceid.upatras.gr heike.adel@de.bosch.com} \\
{\tt inquiries@cislmu.org} \\\And
  Hinrich Sch\"{u}tze$^1$ \\\\\\\\
 }

\date{}
\def\uprm#1{\mbox{$^{\hbox{\scriptsize #1}}$}}
\def\mathlinebreak{\\[0.3cm]}
\def\mathindent{\mbox{\hspace{0.3cm}}}
\begin{document}
\maketitle
\begin{abstract}
Character-level models of tokens have been shown to be 
effective at dealing with within-token noise and out-of-vocabulary words. 
However, they often still rely on correct token boundaries.  
In this paper, we propose 
to eliminate the need for tokenizers
with an end-to-end character-level semi-Markov conditional random field.
It uses neural networks for its character and segment representations.
We demonstrate its effectiveness in multilingual settings
and when token boundaries are noisy:
It matches state-of-the-art part-of-speech taggers for various languages
and significantly outperforms them on a noisy English version of a benchmark dataset.
Our code and the noisy dataset are publicly available at \url{http://cistern.cis.lmu.de/semiCRF}.
\end{abstract}

\section{Introduction}
Recently, character-based neural networks (NNs) gained popularity
for different tasks,
ranging from text classification \cite{zhang2015character} 
and language modeling  \cite{kim2016character} to machine translation \cite{luong2016achieving}. 
Character-level models are attractive since they 
can effectively model morphological variants of words
and build representations even for unknown words, suffering 
less from out-of-vocabulary problems \cite{pinter2017mimicking}.
\blfootnote{* Work was done at Center for Information and Language Processing, LMU Munich.}

However, most character-level models still rely
on tokenization and use characters only for creating more robust
token representations
\cite{santos2014learning, lample2016neural,ma2016end,plank2016multilingual}. 
This leads to high performance on well-formatted text
or text with misspellings \cite{yu2017general, DBLP:conf/aaai/SakaguchiDPD17}
but ties the performance to the quality of the tokenizer.
While humans are very robust to noise caused by insertion of spaces
(e.g., ``car nival'')
or deletion of spaces 
(``deeplearning''),
this can cause severe underperformance of machine learning models.
Similar challenges arise for languages with difficult
tokenization, such as Chinese or Vietnamese.
For text with difficult or noisy tokenization, more robust models are needed.

In order to address this challenge, we propose a model
that does not require any tokenization.
It is based on semi-Markov conditional random fields (semi-CRFs)
\cite{sarawagi2005semi} which jointly learn to segment (tokenize) and label
the input (e.g., characters).
To represent the character segments, we compare
different NN approaches.

In our experiments, we address part-of-speech (POS) tagging.
However,  
our model is generally applicable to other sequence-tagging
tasks as well 
since it does not require any task-specific
hand-crafted features.
Our model achieves state-of-the-art results on the Universal Dependencies dataset \cite{ud12}. 
To demonstrate its effectiveness, we evaluate it not only on English
but also on languages with inherently difficult tokenization, namely
Chinese, Japanese and Vietnamese.
We further analyze the robustness of our model against difficult tokenization by
randomly corrupting the tokenization of the English dataset.
Our model
significantly 
outperforms
state-of-the-art
token-based models 
in this analysis.

Our contributions are: 
1) We present a truly end-to-end character-level sequence tagger 
that does not rely on any tokenization and achieves state-of-the-art
results across languages. 
2) We show its robustness against noise caused by corrupted tokenization, 
further establishing the importance of 
character-level models as a promising research direction.
3) For future research, our code and the noisy version
of the dataset are publicly available at \url{http://cistern.cis.lmu.de/semiCRF}.

\section{Model}
This section describes our model which is also depicted in Figure \ref{fig:ourModel}.

\subsection{Character-based Input Representation}
\label{sec:model-input} 
\begin{figure}
	\centering
\includegraphics[width=.8\columnwidth]{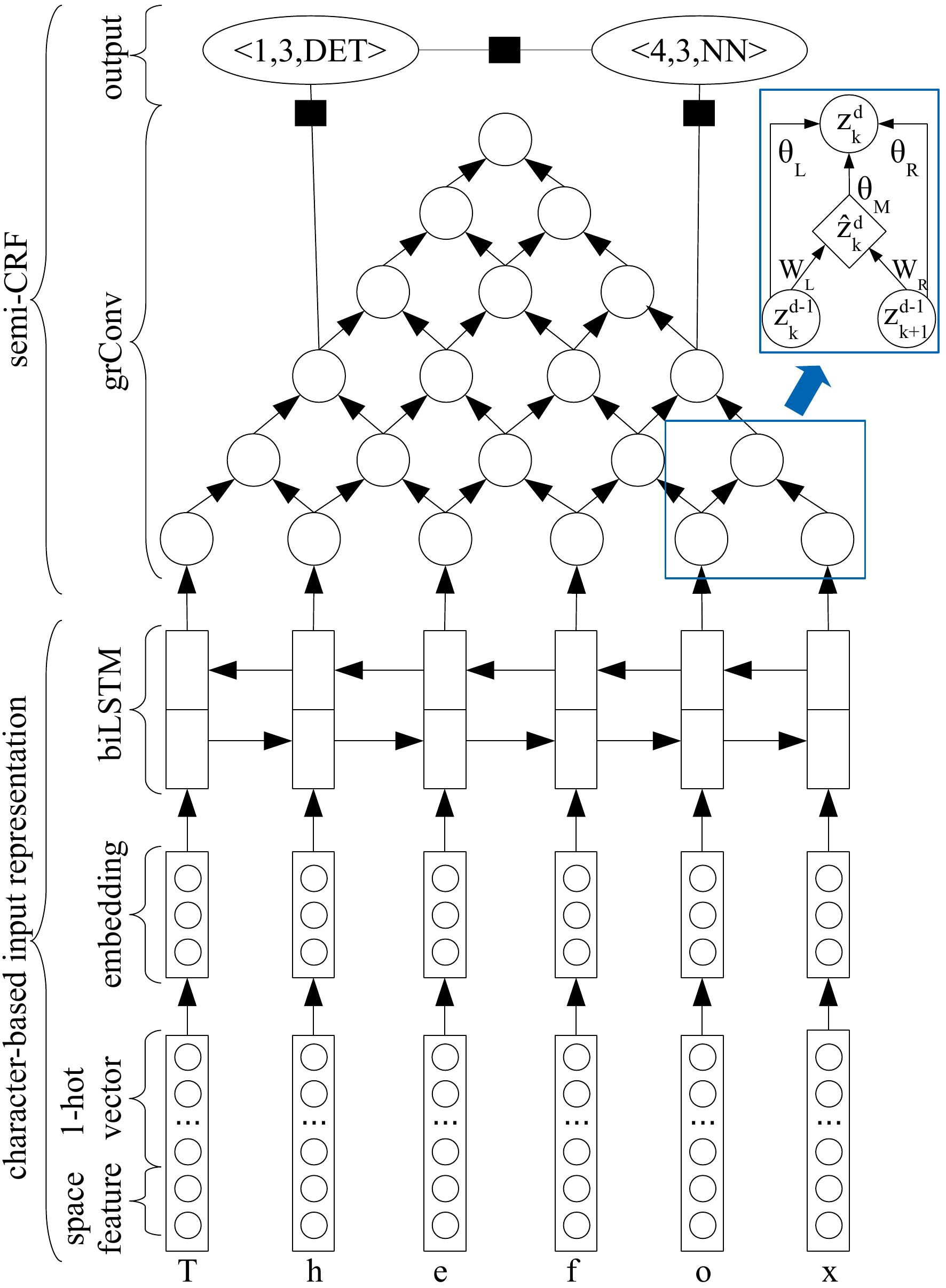}
\caption{Overview of our model. Illustration of gating for grConv taken from \cite{zhuo2016segment}.}
\label{fig:ourModel}
\end{figure}

The input to our model 
is the raw character sequence.
We convert each character 
to a one-hot representation. 
Out-of-vocabulary characters are represented with a zero vector.
Our vocabulary does not include the space character
since there is no part-of-speech label for it.
Instead, our model represents space as two ``space
features'' (lowest level in Figure~\ref{fig:ourModel}):
two binary dimensions indicate whether the previous or
next character is a space.
Then, a linear transformation 
is applied to the extended one-hot encoding to produce a character embedding. 
The character embeddings are fed into a bidirectonal LSTM (biLSTM)
\cite{hochreiter1997long} 
that computes 
context-aware representations. 
These representations form the input to the segment-level feature extractor.

\subsection{Semi-Markov CRF}
Our model partitions a sequence of characters 
$x = $ \{$ x_1,\dots, x_T $\}$ $ of length $T$, 
into (token-like) segments $ s = $ \{$ s_1,\dots, s_{|s|}    $ \}$ $
  with $s_j = \langle a_j, d_j, y_j \rangle$ where $a_j$ 
is the starting position of the $j$\uprm{th} segment, 
$d_j$ is its length and $y_j$ is its label. 
Thus, it assigns the same label $y_j$ to the whole segment $s_j$. 
The sum of the lengths of the segments equals the 
number of non-space characters: $\sum_{j=1}^{|s|} d_j = T$.\footnote{For
efficiency, we define a maximum segment length $L$: $d_j < L, 1 \le j \le |s|$. $L$ is
a hyperparameter. We
choose it based on the observed segment lengths in the training set.}

The semi-CRF defines the conditional distribution of the input segmentations as:\mathlinebreak
\mathindent$p(s|x)\!\!=\!\!\frac{1}{Z(x)}\text{exp}(\sum_{j=1}^{|s|} F(s_j, x) \!\!+\!\! A(y_{j-1} , y_j))$\mathlinebreak
\mathindent$Z(x)\!\!=\!\!\sum_{s'\in S}\text{exp}(\sum_{j=1}^{|s'|}F(s'_j,x)\!\!+\!\!A(y'_{j-1},y'_j))$\mathlinebreak
where $ F(s_j, x)  $ is the score for segment $s_j$ (including its label $y_j$), 
and $A(y_{t-1} , y_t) $ is the transition score of the labels of two adjacent segments.
Thus, $p(s|x)$ jointly models the segmentation and label assignment.
For the normalization term $Z(x)$, we sum over the set of all possible segmentations $S$.

The score  $F(s_j, x)$
is computed as: \mathlinebreak
\mathindent$F(s_j, x)  = \mathbf{w}^\top_{y_j} f(s_j, x) + b_{y_j}$\mathlinebreak
where $W = (\mathbf{w}_1, \ldots,\mathbf{w}_{|Y|})^\top  \in \mathbb{R}^{|Y|\times D}  $ and
$ b = (b_1, \ldots, b_{|Y|})^\top  \in \mathbb{R}^{|Y|}$
are trained parameters,
$ f(s_j, x) \in \mathbb{R}^{D} $ 
is the feature representation of the labeled segment $s_j$, $|Y|$ is the number of 
output classes and $D$ is the length of the segment representation. 

For training and decoding, we use the 
semi-Markov analogies of the forward and Viterbi algorithm, respectively \cite{sarawagi2005semi}. 
In order to avoid numerical instability, 
all computations are performed in log-space.

\subsubsection{Segment-level Features}
\label{sec:grConv}
\newcite{sarawagi2005semi} and \newcite{yang2012extracting}
compute segment-level features
by hand-crafted rules.
Recent work learns the features automatically
with NNs \cite{kong2015segmental,zhuo2016segment}. 
This avoids the manual design of new features for new languages/tasks.
We adopt
\textbf{\textit{Gated Recursive Convolutional Neural Networks} (grConv)} 
\cite{cho2014properties,zhuo2016segment}
since they allow to hierarchically combine features for segments.
We argue that this is especially useful for compositionality in language.
An example is the word ``airport'' which can be composed of the segments
``air'' and ``port''.

GrConv constructs features 
by recursively combining adjacent segment 
representations
in a pyramid shape way (see Figure \ref{fig:ourModel}). 
The $d$\uprm{th} level of the pyramid 
consists of all representations for segments of length $d$. 
The first level holds
the character representations from our biLSTM. 
The representation $z_k^{(d)} \in \mathbb{R}^D$, stored in the $k$\uprm{th} node of layer $d$,
is computed as follows:\mathlinebreak
\mathindent$z_{k}^{(d)} = \theta_L \circ z_{k}^{(d-1)} + \theta_R \circ z_{k+1}^{(d-1)} + \theta_M \circ \hat{z_{k}}^{(d)}$\mathlinebreak
\mathindent$\text{with } \hat{z_{k}}^{(d)} = g(W_L z_{k}^{(d-1)} + W_R z_{k+1}^{(d-1)} + b_w)$\mathlinebreak
where $W_L, W_R \in \mathbb{R}^{D \times D}$ and $b_w \in \mathbb{R}^{D}$ are 
globally shared parameters, $\theta_L$, $\theta_M$ and $\theta_R$ are gates,
$g$ is a non-linearity
and $\circ$ denotes element-wise multiplication. The gates are illustrated
in the blue box of Figure \ref{fig:ourModel} and described in \cite{zhuo2016segment}.

\section{Experiments and Analysis}
Our implementation is in PyTorch
\cite{paszke2017automatic}.  Hyperparameters are tuned
on the development set.
We use mini-batch gradient descent with a batch size of 20
and Adam \cite{kingma2014adam} as the optimizer.
The learning rate is 1e-3, the coefficients for computing running
averages of the gradient and its square are 0.9 and 0.999, respectively.
A term of 1e-8 is added to the denominator for numerical stability.
We use character embeddings of size 60 and three stacked 
biLSTM layers with 100 hidden units for each direction. For the semi-CRF, we set the 
maximum segment length to $L=23$ as tokens of bigger length are rarely seen in the training sets.
To avoid overfitting, we apply dropout with a probability of 0.25 on each layer 
including the input. For input dropout, we randomly replace a character embedding
with a zero vector, similar to \newcite{gillick2015multilingual}. This avoids
overfitting to local character patterns. Moreover, we employ early stopping on the development set
with a minimum of 20 training epochs.
We run our experiments on a gpu which speeds up the training compared to multiple cpu cores
considerably. We assume that it especially benefits from parallelizing the computation 
of each level of the grConv pyramid.

\subsection{Multilingual Experiments on Clean Data}
\textbf{Data and Evaluation.} 
To compare our model
to state-of-the-art character-based POS taggers, we evaluate 
its accuracy on the
English part of the Universal Dependencies (UD) v1.2 
dataset \cite{ud12}.
For multilingual experiments,
we use the English (EN), Chinese (ZH), Japanese (JA) and Vietnamese (VI) 
part of UD v2.0\footnote{UD v1.2 
does not provide data for JA, VI, ZH.} \cite{nivre2017universal},
using the splits, training and evaluation rules 
from the CoNNL 2017 shared task
\cite{zeman2017conll}. In particular, we calculate joint tokenization and
UPOS (universal POS) $F_1$ scores.

\textbf{Baselines for UD v1.2.}
We compare our model to two character-based models
that are state of the art on UD v1.2:
\textbf{bilstm-aux} \cite{plank2016multilingual} and 
\textbf{CNN Tagger} \cite{yu2017general}. 
We also compare to a state-of-the-art word-based CRF model 
\textbf{MarMot}\footnote{\url{http://cistern.cis.lmu.de/marmot/}}
\cite{muller2015robust}.

\begin{table}
\footnotesize
\centering
\begin{tabular}{c | c c c }
Model  & $\vec{w}$ & $\vec{c}$   \\\hline
MarMot & \textbf{94.36}  & -  \\
bilstm-aux & 92.10  & 91.62   \\
CNN Tagger  & 92.64  & 93.76   \\
\hline
Our & -  & \textbf{94.27}   \\
Our without space feature & - & 93.35\\
Our with SRNN & -  & 93.86  \\
\end{tabular}
\caption{POS tag accuracy on UD v1.2 (EN). \\'-' denotes that the model does not use this input.}
\label{tab:resultsClean}
\end{table}

\textbf{Results on English (UD v1.2).}
Table \ref{tab:resultsClean} provides our results on UD v1.2,
categorizing the models into token-level 
($\vec{w}$) and character-only models ($\vec{c}$). 
While most pure character-level models
cannot ensure consistent labels for each character of a token, 
our semi-CRF outputs correct segments in most cases (tokenization $F_1$ is 98.69\%, see Table \ref{tab:corr}), 
and ensures a single label for all characters of a segment. 
Our model achieves the best results among
all character-level models and comparable results to the word-level model MarMot. 

\begin{table*}
\footnotesize
\centering
\begin{tabular}{c | c c | c c | cc | cc | cc| c c }
 & \multicolumn{2}{c|}{UDPipe 1.2}  & \multicolumn{2}{c|}{Stanford} & \multicolumn{2}{c|}{FBAML} & \multicolumn{2}{c|}{TRL} & \multicolumn{2}{c|}{IMS} &  \multicolumn{2}{c}{Our}\\
  & Tokens & POS   & Tokens & POS  & Tokens & POS & Tokens & POS & Tokens & POS & Tokens & POS\\\hline
EN & \textbf{99.03}  & 93.50 & 98.67  & \textbf{95.11}  & \underline{98.98} & \underline{94.09} & 94.31 & 82.41 & 98.67 & 93.29 & 98.79  & 93.45 \\
JA & 90.97  & 88.19 & 89.68  & 88.14  &  93.32 & 91.04 & \textbf{98.59} & \textbf{98.45} & 91.68 & 89.07 & \underline{93.86}  & \underline{91.34} \\
VI & 84.26  & 75.29 & 82.47  & 75.28  & 83.80 & 75.84 & 85.41 & 74.53 & \underline{86.67} & \textbf{77.88} &\textbf{88.06}  & \underline{77.67}\\
ZH  &  89.55   & 83.47  & 88.91  & 85.26 & \textbf{94.57} & \textbf{88.36} & 83.64 & 71.31 & 92.81 & 86.33 & \underline{93.82}  & \underline{88.15} \\
\hline
Avg & 90.95 & 85.11 & 89.93 & 85.95 & \underline{92.67} & \underline{87.33} & 90.49 & 81.68 & 92.46 & 86.64 & \textbf{93.66} & \textbf{87.65}
\end{tabular}
\caption{Tokenization and joint token-POS $F_1$ on UD v2.0. Best scores are in bold, second-best are underlined.}
\label{tab:resultsMulti2}
\end{table*}

In addition, we assess the impact of two components of our model:
the space feature (see Section \ref{sec:model-input}) and
grConv (see Section \ref{sec:grConv}).
Table~\ref{tab:resultsClean} shows that the performance of our model
decreases when ablating the space feature, confirming that information about spaces
plays a valuable role for English.
To evaluate the effectiveness of grConv for segment representations,
we replace it with a 
\textit{\textbf{Segmental Recurrent Neural Network (SRNN)}} 
\cite{kong2015segmental}.\footnote{In an initial experiment, we also
replaced it with a simpler method that creates a segment representation
by subtracting the character biLSTM hidden state of the segment start
from the hidden state of the segment end. This is one of the segment-level features
employed, for instance, by \newcite{ye2018}. However, this approach did not lead to promising
results in our case. We assume that more sophisticated methods like grConv or SRNN are needed
in this setup.}
SRNN uses dynamic programming and biLSTMs to create segment representations.
Its performance is slightly worse compared to grConv (last row of Table \ref{tab:resultsClean}).
We attribute this to the different way of feature creation: 
While grConv 
hierarchically combines context-enhanced n-grams, SRNN 
constructs segments in a sequential order. The latter may
be less suited for compositional segments like
``airport''.

\textbf{Baselines for UD v2.0.}
We compare to the top performing models for 
EN, JA, VI, ZH from the CoNLL 2017 shared task:
UDPipe 1.2 \cite{straka2017tokenizing}, 
Stanford \cite{dozat2017stanford},
FBAML \cite{qian-liu2017},
TRL \cite{kanayama2017}, and
IMS \cite{bjorkelund2017}.

\textbf{Multilingual Results (UD v2.0).}
Table \ref{tab:resultsMulti2} provides our results.
While for each language another shared task system performs best, our system
performs consistently well across languages (best or
second-best except for EN),
leading to the best average scores
for both tokenization and POS tagging. Moreover, it matches the state of the art 
for Chinese (ZH) and Vietnamese (VI), two languages with very different characteristics in tokenization.

\subsection{Analysis on Noisy Data}
To further investigate the robustness of our model, we conduct experiments with
different levels of corrupted tokenization in English.
We argue that this could also give us insights into why it performs
well on languages with difficult tokenization, e.g., on Chinese
which omits spaces between tokens, or on Vietnamese which has spaces inside
tokens, after each syllable.
Note that we do not 
apply input dropout for these experiments, 
since the corrupt tokenization already acts as a regularizer.

\textbf{Data.} 
We are not aware of a POS tagging dataset with corrupted tokenization.
Thus, we create one based on UD v1.2 (EN).
For each token, we either delete the space after it with probability 
$P = p_d$ or insert a space between 
two characters with $P = p_i$:
\textit{"The fox chased the rabbit"}
$\rightarrow$ \textit{"The f ox cha sed therabbit"}. 
We vary $p_d$ and $p_i$ to construct three datasets
with different noise levels (LOW, MID, HIGH, see Table \ref{tab:statNoisy}).
We note that there are more sophisticated ways
of creating ``errors'' in text. An example is \newcite{D18-1541}
who generate grammatical errors. We leave
the investigation of other methods for generating tokenization errors to future work.

\begin{table}[h]
\centering
\footnotesize
\begin{tabular}{l|cc|cc}
\toprule
level & $p_d$ & \# deletions & $p_i$ & \# insertions\\\hline
LOW & 0.1 & 15198 & 0.05 & 26497\\
MID & 0.3 & 39361 & 0.11 & 40474\\
HIGH & 0.6 & 65387 & 0.33 & 68209\\ \bottomrule
\end{tabular}
\caption{Noisy dataset statistics (three different noise levels).}
\label{tab:statNoisy}
\end{table}

\begin{figure}[h]
\centering
\includegraphics[width=.4\textwidth]{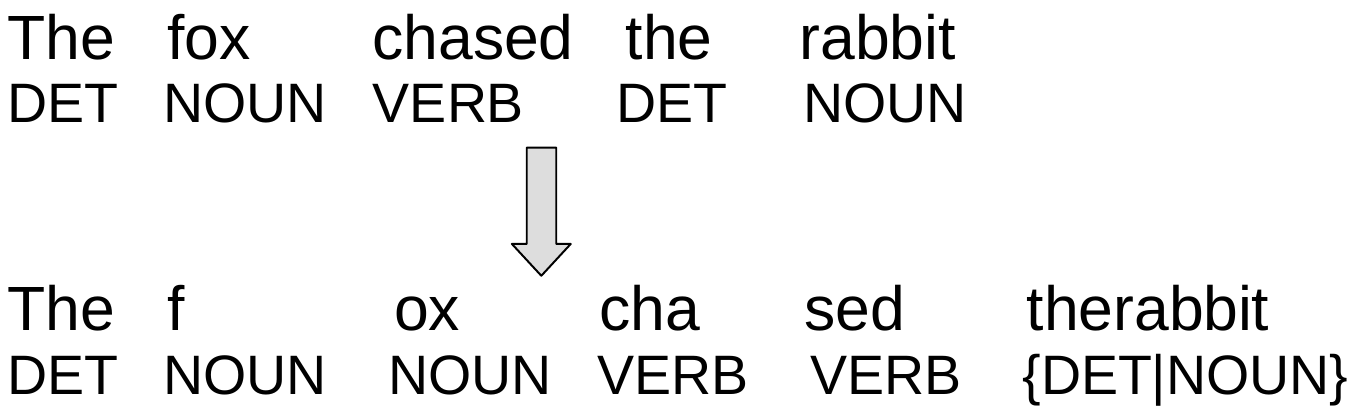}
\caption{Example of label assignment.}
\label{fig:labels}
\end{figure}

\textbf{Labeling.}
As mentioned before, we either delete the space after a token with probability 
$p_d$ or insert a space between 
two of its characters with probability $p_i$.
We assign the label from the original token to every sub-token 
created by space insertion. For space deletions, we
randomly choose one of the two original labels for training and evaluate
against the union of them.
Figure \ref{fig:labels} shows an example.

\textbf{Baseline.}
We compare our joint model to a traditional pipeline of 
tokenizer (UDpipe 1.0)\footnote{http://lindat.mff.cuni.cz/services/udpipe/}
and token-level POS tagger (MarMot).\footnote{In contrast to Table \ref{tab:resultsClean} where we use gold tokens for MarMot.}
We re-train MarMot on the corrupted datasets.

\textbf{Evaluation.}
We evaluate the models on the noisy datasets using two different metrics:
(i) tokenization and joint token-POS $F_1$ as in Table \ref{tab:resultsMulti2},
and (ii) a relaxed variant of POS tag accuracies.
With the latter, we can assess the performance of MarMot without
penalizing it for potential errors of UDpipe.
For calculating the relaxed accuracy, we count the POS tag of a gold
token as correct if MarMot predicts the tag for any subpart of it.
We provide more details on the relaxed evaluation (description, examples and implementation) 
in our code repository.
Note that we apply the relaxed evaluation only to UDpipe+MarMot
but not to our model. The output of our model is directly evaluated against the gold
labels of the clean corpus.

\textbf{Results.}
The
performance of our model
decreases only slightly when increasing the noise level while the performance of
UDpipe+MarMot drops significantly (Table \ref{tab:corr}). 
This confirms that our model is robust against
noise from tokenization. Note that most other character-based models
would suffer from the same performance drop as MarMot
since they rely on tokenized inputs.

\begin{table}
\footnotesize
\centering
\begin{tabular}{p{.9cm} | p{.6cm}p{.6cm}|p{.6cm} | p{.6cm}p{.6cm}|p{.6cm}}
	   & \multicolumn{3}{c|}{UDpipe+MarMot} & \multicolumn{3}{c}{Our} \\
& \multicolumn{2}{c|}{$F_1$} & acc & \multicolumn{2}{c|}{$F_1$} & acc\\
           Noise      & Tokens & POS & POS & Tokens & POS & POS\\
\hline
	CLEAN & 98.48 & 92.75 & 93.48 & \textbf{98.69} & \textbf{93.48}  & \textbf{94.27}\\
\hline
	LOW & 70.90 & 65.56 & 83.73 & \textbf{96.08} & \textbf{90.51} &\textbf{92.80}\\
	MID  & 20.62 & 19.07 & 58.53 & \textbf{95.28} & \textbf{89.80} &\textbf{92.54}\\
	HIGH   & 20.47 & 18.05 & 56.96 & \textbf{95.45} & \textbf{89.82} & \textbf{92.14}\\
\end{tabular}
\caption{Tokenization $F_1$, joint token-POS $F_1$ and (relaxed) POS tag accuracies on noisy version of UD v1.2.}
\label{tab:corr}
\end{table}

\textbf{Discussion.}
The results in Table \ref{tab:corr} show that our model can reliably recover token boundaries,
even in noisy scenarios.
This also explains its strong performance across languages:
It can handle different languages,
independent of whether the language merges
tokens without whitespaces (e.g., Chinese) or separates
tokens with whitespaces into syllables (e.g., Vietnamese).

\section{Related Work}
\textbf{Character-based POS Tagging.}
Most work 
uses characters only to build more robust token representations 
but
still relies on external tokenizers 
\cite{santos2014learning,lample2016neural,plank2016multilingual,dozat2017stanford, liu2017empower}. 
In contrast, our model jointly learns segmentation and POS tagging.
\newcite{gillick2015multilingual} do not rely on tokenization either
but in contrast to their greedy decoder, our model optimizes the whole
output sequence and is 
able to
revise local decisions \cite{Lafferty2001}.
For 
processing characters,
LSTMs 
\cite{lample2016neural,plank2016multilingual,dozat2017stanford} or CNNs \cite{ma2016end,yu2017general} are used.
Our model combines biLSTMs and
grConv to model both the context of characters (LSTM)
and the compositionality of language (grConv).

\textbf{Joint Segmentation and POS Tagging.}
The
top performing models
of EN, JA, VI and ZH
use a pipeline of tokenizer and word-based POS tagger
but do not treat both tasks jointly 
\cite{bjorkelund2017,dozat2017stanford,kanayama2017,qian-liu2017}.
Especially for Chinese, there is a lot of work
on joint word segmentation and POS tagging, e.g.,
\cite{zhang-clark2008,sun2011,hatori2012,zheng2013,kong2015segmental,cai2016neural,chen2017,shao2017},
of which some use CRFs to predict one POS tag per character.
However, this is hard to transfer to languages
like English and Vietnamese where single characters are less informative
and tokens are much longer, resulting in a larger combinatory label space.
Thus, we choose a semi-Markov formalization to directly model
segments.

\textbf{Semi-Markov CRFs for Sequence Tagging.}
\newcite{zhuo2016segment} and \newcite{ye2018}
apply semi-CRFs to word-level inputs 
for named entity recognition. 
In contrast, we model character-based POS tagging.
Thus, the expected length of our character segments is considerably
larger than the expected length of word-based segments for NER.
\newcite{kong2015segmental}
build SRNNs that we use as a baseline.
In contrast to their 0-order model, we train a 1-order semi-CRF
to model
dependencies between segment labels.

\section{Conclusion}
We presented an end-to-end model for character-based
part-of-speech tagging that uses
semi-Markov conditional random fields to jointly
segment and label a sequence of characters.
Input representations and  segment representations
are trained parameters learned in end-to-end training by the
neural network part of the model.
The model achieves state-of-the-art results on two benchmark datasets
across several typologically diverse languages.
By corrupting the tokenization
of the dataset, we show the robustness of our model, explaining
its good performance on languages with difficult tokenization.

\section*{Acknowledgments}
This work was funded by the European Research Council (ERC \#740516).
We would like to thank the anonymous reviewers
for their helpful comments.

\bibliography{refs}

\begin{thebibliography}{39}
\expandafter\ifx\csname natexlab\endcsname\relax\def\natexlab#1{#1}\fi

\bibitem[{Bj\"{o}rkelund et~al.(2017)Bj\"{o}rkelund, Falenska, Yu, and
  Kuhn}]{bjorkelund2017}
Anders Bj\"{o}rkelund, Agnieszka Falenska, Xiang Yu, and Jonas Kuhn. 2017.
\newblock {IMS} at the {CoNLL} 2017 {UD} shared task: {CRFs} and perceptrons
  meet neural networks.
\newblock In \emph{Proceedings of the CoNLL 2017 Shared Task: Multilingual
  Parsing from Raw Text to Universal Dependencies}, pages 40--51, Vancouver,
  Canada. Association for Computational Linguistics.

\bibitem[{Cai and Zhao(2016)}]{cai2016neural}
Deng Cai and Hai Zhao. 2016.
\newblock Neural word segmentation learning for chinese.
\newblock In \emph{Proceedings of the 54th Annual Meeting of the Association
  for Computational Linguistics (Volume 1: Long Papers)}, pages 409--420,
  Berlin, Germany. Association for Computational Linguistics.

\bibitem[{Chen et~al.(2017)Chen, Qiu, and Huang}]{chen2017}
Xinchi Chen, Xipeng Qiu, and Xuanjing Huang. 2017.
\newblock A feature-enriched neural model for joint chinese word segmentation
  and part-of-speech tagging.
\newblock In \emph{Proceedings of the 26th International Joint Conference on
  Artificial Intelligence}, pages 3960--3966, Melbourne, Australia. AAAI Press.

\bibitem[{Cho et~al.(2014)Cho, van Merrienboer, Bahdanau, and
  Bengio}]{cho2014properties}
Kyunghyun Cho, Bart van Merrienboer, Dzmitry Bahdanau, and Yoshua Bengio. 2014.
\newblock On the properties of neural machine translation: Encoder--decoder
  approaches.
\newblock In \emph{Proceedings of SSST-8, Eighth Workshop on Syntax, Semantics
  and Structure in Statistical Translation}, pages 103--111, Doha, Qatar.
  Association for Computational Linguistics.

\bibitem[{Dozat et~al.(2017)Dozat, Qi, and Manning}]{dozat2017stanford}
Timothy Dozat, Peng Qi, and Christopher~D Manning. 2017.
\newblock Stanford's graph-based neural dependency parser at the {CoNLL} 2017
  shared task.
\newblock In \emph{Proceedings of the CoNLL 2017 Shared Task: Multilingual
  Parsing from Raw Text to Universal Dependencies}, pages 20--30, Vancouver,
  Canada. Association for Computational Linguistics.

\bibitem[{Gillick et~al.(2016)Gillick, Brunk, Vinyals, and
  Subramanya}]{gillick2015multilingual}
Dan Gillick, Cliff Brunk, Oriol Vinyals, and Amarnag Subramanya. 2016.
\newblock Multilingual language processing from bytes.
\newblock In \emph{Proceedings of the 2016 Conference of the North American
  Chapter of the Association for Computational Linguistics: Human Language
  Technologies}, pages 1296--1306, San Diego, California. Association for
  Computational Linguistics.

\bibitem[{Hatori et~al.(2012)Hatori, Matsuzaki, Miyao, and Tsujii}]{hatori2012}
Jun Hatori, Takuya Matsuzaki, Yusuke Miyao, and Jun'ichi Tsujii. 2012.
\newblock Incremental joint approach to word segmentation, pos tagging, and
  dependency parsing in chinese.
\newblock In \emph{Proceedings of the 50th Annual Meeting of the Association
  for Computational Linguistics (Volume 1: Long Papers)}, pages 1045--1053,
  Jeju Island, Korea. Association for Computational Linguistics.

\bibitem[{Hochreiter and Schmidhuber(1997)}]{hochreiter1997long}
Sepp Hochreiter and J{\"u}rgen Schmidhuber. 1997.
\newblock Long short-term memory.
\newblock \emph{Neural computation}, 9(8):1735--1780.

\bibitem[{Kanayama et~al.(2017)Kanayama, Muraoka, and Yoshikawa}]{kanayama2017}
Hiroshi Kanayama, Masayasu Muraoka, and Katsumasa Yoshikawa. 2017.
\newblock A semi-universal pipelined approach to the conll 2017 ud shared task.
\newblock In \emph{Proceedings of the {CoNLL} 2017 Shared Task: Multilingual
  Parsing from Raw Text to Universal Dependencies}, pages 265--273, Vancouver,
  Canada. Association for Computational Linguistics.

\bibitem[{Kasewa et~al.(2018)Kasewa, Stenetorp, and Riedel}]{D18-1541}
Sudhanshu Kasewa, Pontus Stenetorp, and Sebastian Riedel. 2018.
\newblock \href {http://aclweb.org/anthology/D18-1541} {Wronging a right:
  Generating better errors to improve grammatical error detection}.
\newblock In \emph{Proceedings of the 2018 Conference on Empirical Methods in
  Natural Language Processing}, pages 4977--4983, Brussels, Belgium.
  Association for Computational Linguistics.

\bibitem[{Kim et~al.(2016)Kim, Jernite, Sontag, and Rush}]{kim2016character}
Yoon Kim, Yacine Jernite, David Sontag, and Alexander~M Rush. 2016.
\newblock Character-aware neural language models.
\newblock In \emph{{AAAI} Conference on Artificial Intelligence}, pages
  2741--2749. AAAI Press.

\bibitem[{Kingma and Ba(2014)}]{kingma2014adam}
Diederik~P Kingma and Jimmy Ba. 2014.
\newblock Adam: A method for stochastic optimization.
\newblock In \emph{International Conference on Learning Representations}.

\bibitem[{Kong et~al.(2015)Kong, Dyer, and Smith}]{kong2015segmental}
Lingpeng Kong, Chris Dyer, and Noah~A Smith. 2015.
\newblock Segmental recurrent neural networks.
\newblock \emph{arXiv preprint arXiv:1511.06018}.

\bibitem[{Lafferty et~al.(2001)Lafferty, McCallum, and Pereira}]{Lafferty2001}
John~D. Lafferty, Andrew McCallum, and Fernando C.~N. Pereira. 2001.
\newblock Conditional random fields: Probabilistic models for segmenting and
  labeling sequence data.
\newblock In \emph{International Conference on Machine Learning}, pages
  282--289. Morgan Kaufmann Publishers Inc.

\bibitem[{Lample et~al.(2016)Lample, Ballesteros, Subramanian, Kawakami, and
  Dyer}]{lample2016neural}
Guillaume Lample, Miguel Ballesteros, Sandeep Subramanian, Kazuya Kawakami, and
  Chris Dyer. 2016.
\newblock Neural architectures for named entity recognition.
\newblock In \emph{Proceedings of the 2016 Conference of the North American
  Chapter of the Association for Computational Linguistics: Human Language
  Technologies}, pages 260--270, San Diego, California. Association for
  Computational Linguistics.

\bibitem[{Liu et~al.(2017)Liu, Shang, Xu, Ren, Gui, Peng, and
  Han}]{liu2017empower}
Liyuan Liu, Jingbo Shang, Frank Xu, Xiang Ren, Huan Gui, Jian Peng, and Jiawei
  Han. 2017.
\newblock Empower sequence labeling with task-aware neural language model.
\newblock \emph{arXiv preprint arXiv:1709.04109}.

\bibitem[{Luong and Manning(2016)}]{luong2016achieving}
Minh-Thang Luong and Christopher~D. Manning. 2016.
\newblock Achieving open vocabulary neural machine translation with hybrid
  word-character models.
\newblock In \emph{Proceedings of the 54th Annual Meeting of the Association
  for Computational Linguistics (Volume 1: Long Papers)}, pages 1054--1063,
  Berlin, Germany. Association for Computational Linguistics.

\bibitem[{Ma and Hovy(2016)}]{ma2016end}
Xuezhe Ma and Eduard Hovy. 2016.
\newblock End-to-end sequence labeling via bi-directional {LSTM-CNNs-CRF}.
\newblock In \emph{Proceedings of the 54th Annual Meeting of the Association
  for Computational Linguistics (Volume 1: Long Papers)}, pages 1064--1074,
  Berlin, Germany. Association for Computational Linguistics.

\bibitem[{M\"{u}ller and Sch\"{u}tze(2015)}]{muller2015robust}
Thomas M\"{u}ller and Hinrich Sch\"{u}tze. 2015.
\newblock Robust morphological tagging with word representations.
\newblock In \emph{Proceedings of the 2015 Conference of the North American
  Chapter of the Association for Computational Linguistics: Human Language
  Technologies}, pages 526--536, Denver, Colorado. Association for
  Computational Linguistics.

\bibitem[{Nivre and \v{Z}eljko Agic(2017)}]{nivre2017universal}
Joakim Nivre and Lars~Ahrenberg \v{Z}eljko Agic. 2017.
\newblock Universal dependencies 2.0 {CoNLL} 2017 shared task development and
  test data. lindat/clarin digital library at the institute of formal and
  applied linguistics, charles university.

\bibitem[{Nivre et~al.(2015)}]{ud12}
Joakim Nivre et~al. 2015.
\newblock Universal dependencies 1.2.
\newblock {LINDAT}/{CLARIN} digital library at the Institute of Formal and
  Applied Linguistics ({{\\'U}FAL}), Faculty of Mathematics and Physics,
  Charles University.

\bibitem[{Paszke et~al.(2017)Paszke, Gross, Chintala, Chanan, Yang, DeVito,
  Lin, Desmaison, Antiga, and Lerer}]{paszke2017automatic}
Adam Paszke, Sam Gross, Soumith Chintala, Gregory Chanan, Edward Yang, Zachary
  DeVito, Zeming Lin, Alban Desmaison, Luca Antiga, and Adam Lerer. 2017.
\newblock Automatic differentiation in pytorch.
\newblock In \emph{The future of gradient-based machine learning software and
  techniques, {NIPS} 2017}.

\bibitem[{Pinter et~al.(2017)Pinter, Guthrie, and
  Eisenstein}]{pinter2017mimicking}
Yuval Pinter, Robert Guthrie, and Jacob Eisenstein. 2017.
\newblock Mimicking word embeddings using subword rnns.
\newblock \emph{arXiv preprint arXiv:1707.06961}.

\bibitem[{Plank et~al.(2016)Plank, S{\o}gaard, and
  Goldberg}]{plank2016multilingual}
Barbara Plank, Anders S{\o}gaard, and Yoav Goldberg. 2016.
\newblock Multilingual part-of-speech tagging with bidirectional long
  short-term memory models and auxiliary loss.
\newblock In \emph{Proceedings of the 54th Annual Meeting of the Association
  for Computational Linguistics (Volume 2: Short Papers)}, pages 412--418,
  Berlin, Germany. Association for Computational Linguistics.

\bibitem[{Qian and Liu(2017)}]{qian-liu2017}
Xian Qian and Yang Liu. 2017.
\newblock A non-{DNN} feature engineering approach to dependency parsing --
  {FBAML} at {CoNLL} 2017 shared task.
\newblock In \emph{Proceedings of the CoNLL 2017 Shared Task: Multilingual
  Parsing from Raw Text to Universal Dependencies}, pages 143--151, Vancouver,
  Canada. Association for Computational Linguistics.

\bibitem[{Sakaguchi et~al.(2017)Sakaguchi, Duh, Post, and
  Durme}]{DBLP:conf/aaai/SakaguchiDPD17}
Keisuke Sakaguchi, Kevin Duh, Matt Post, and Benjamin~Van Durme. 2017.
\newblock \href {http://aaai.org/ocs/index.php/AAAI/AAAI17/paper/view/14332}
  {Robsut wrod reocginiton via semi-character recurrent neural network}.
\newblock In \emph{Proceedings of the Thirty-First {AAAI} Conference on
  Artificial Intelligence, February 4-9, 2017, San Francisco, California,
  {USA.}}, pages 3281--3287. {AAAI} Press.

\bibitem[{Santos and Zadrozny(2014)}]{santos2014learning}
C\'{i}cero N.~dos Santos and Bianca Zadrozny. 2014.
\newblock Learning character-level representations for part-of-speech tagging.
\newblock In \emph{International Conference on Machine Learning}, pages
  1818--1826.

\bibitem[{Sarawagi and Cohen(2005)}]{sarawagi2005semi}
Sunita Sarawagi and William~W Cohen. 2005.
\newblock Semi-markov conditional random fields for information extraction.
\newblock In \emph{Advances in Neural Information Processing Systems}, pages
  1185--1192.

\bibitem[{Shao et~al.(2017)Shao, Hardmeier, Tiedemann, and Nivre}]{shao2017}
Yan Shao, Christian Hardmeier, J\"{o}rg Tiedemann, and Joakim Nivre. 2017.
\newblock Character-based joint segmentation and pos tagging for chinese using
  bidirectional rnn-crf.
\newblock In \emph{Proceedings of the Eighth International Joint Conference on
  Natural Language Processing (Volume 1: Long Papers)}, pages 173--183, Taipei,
  Taiwan. Asian Federation of Natural Language Processing.

\bibitem[{Straka and Strakov{\'a}(2017)}]{straka2017tokenizing}
Milan Straka and Jana Strakov{\'a}. 2017.
\newblock Tokenizing, pos tagging, lemmatizing and parsing {UD} 2.0 with
  {UDPipe}.
\newblock In \emph{Proceedings of the CoNLL 2017 Shared Task: Multilingual
  Parsing from Raw Text to Universal Dependencies}, pages 88--99, Vancouver,
  Canada. Association for Computational Linguistics.

\bibitem[{Sun(2011)}]{sun2011}
Weiwei Sun. 2011.
\newblock A stacked sub-word model for joint chinese word segmentation and
  part-of-speech tagging.
\newblock In \emph{Proceedings of the 49th Annual Meeting of the Association
  for Computational Linguistics: Human Language Technologies}, pages
  1385--1394, Portland, Oregon, USA. Association for Computational Linguistics.

\bibitem[{Yang and Cardie(2012)}]{yang2012extracting}
Bishan Yang and Claire Cardie. 2012.
\newblock Extracting opinion expressions with semi-markov conditional random
  fields.
\newblock In \emph{Proceedings of the 2012 Joint Conference on Empirical
  Methods in Natural Language Processing and Computational Natural Language
  Learning}, pages 1335--1345, Jeju Island, Korea. Association for
  Computational Linguistics.

\bibitem[{Ye and Ling(2018)}]{ye2018}
Zhi-Xiu Ye and Zhen-Hua Ling. 2018.
\newblock Hybrid semi-markov crf for neural sequence labeling.
\newblock In \emph{Proceedings of the 56th Annual Meeting of the Association
  for Computational Linguistics (Volume 2: Short Papers)}, Melbourne,
  Australia. Association for Computational Linguistics.

\bibitem[{Yu et~al.(2017)Yu, Falenska, and Vu}]{yu2017general}
Xiang Yu, Agnieszka Falenska, and Ngoc~Thang Vu. 2017.
\newblock A general-purpose tagger with convolutional neural networks.
\newblock In \emph{Proceedings of the First Workshop on Subword and Character
  Level Models in NLP}, pages 124--129, Copenhagen, Denmark. Association for
  Computational Linguistics.

\bibitem[{Zeman et~al.(2017)Zeman, Popel, Straka, Hajic, Nivre, Ginter,
  Luotolahti, Pyysalo, Petrov, Potthast et~al.}]{zeman2017conll}
Daniel Zeman, Martin Popel, Milan Straka, Jan Hajic, Joakim Nivre, Filip
  Ginter, Juhani Luotolahti, Sampo Pyysalo, Slav Petrov, Martin Potthast,
  et~al. 2017.
\newblock Conll 2017 shared task: multilingual parsing from raw text to
  universal dependencies.
\newblock \emph{Proceedings of the CoNLL 2017 Shared Task: Multilingual Parsing
  from Raw Text to Universal Dependencies}, pages 1--19.

\bibitem[{Zhang et~al.(2015)Zhang, Zhao, and LeCun}]{zhang2015character}
Xiang Zhang, Junbo Zhao, and Yann LeCun. 2015.
\newblock Character-level convolutional networks for text classification.
\newblock In \emph{Advances in Neural Information Processing Systems}, pages
  649--657.

\bibitem[{Zhang and Clark(2008)}]{zhang-clark2008}
Yue Zhang and Stephen Clark. 2008.
\newblock Joint word segmentation and {POS} tagging using a single perceptron.
\newblock In \emph{Proceedings of ACL-08: HLT}, pages 888--896, Columbus, Ohio.
  Association for Computational Linguistics.

\bibitem[{Zheng et~al.(2013)Zheng, Chen, and Xu}]{zheng2013}
Xiaoqing Zheng, Hanyang Chen, and Tianyu Xu. 2013.
\newblock Deep learning for {Chinese} word segmentation and {POS} tagging.
\newblock In \emph{Proceedings of the 2013 Conference on Empirical Methods in
  Natural Language Processing}, pages 647--657, Seattle, Washington, USA.
  Association for Computational Linguistics.

\bibitem[{Zhuo et~al.(2016)Zhuo, Cao, Zhu, Zhang, and Nie}]{zhuo2016segment}
Jingwei Zhuo, Yong Cao, Jun Zhu, Bo~Zhang, and Zaiqing Nie. 2016.
\newblock Segment-level sequence modeling using gated recursive semi-markov
  conditional random fields.
\newblock In \emph{Proceedings of the 54th Annual Meeting of the Association
  for Computational Linguistics (Volume 1: Long Papers)}, pages 1413--1423,
  Berlin, Germany. Association for Computational Linguistics.

\end{thebibliography}
\bibliographystyle{acl_natbib}

\end{document}